\documentclass[10pt,a4paper]{article}
\usepackage{graphicx}
\usepackage{multirow}
\usepackage{amsmath}
\usepackage{amssymb}
\usepackage{mathtools}
\usepackage{booktabs}
\usepackage{algorithm}
\usepackage{algorithmic}
\usepackage{tikz}
\usepackage{array}
\usepackage{tabularx}
\usepackage{xcolor}
\usepackage[numbers]{natbib}
\usepackage{caption}
\usepackage{subcaption}
\usepackage{authblk} % Package for author affiliations
\usepackage{geometry}
\usepackage{hyperref}

\usepackage{times}
\usepackage{newtxtext,newtxmath}

\usepackage{setspace}
\begin{document}

\title{\Large\textbf{Cyberbullying Detection in Hinglish Text Using MURIL and Explainable AI}}

\author[1]{Devesh Kumar\thanks{23mcs112@nith.ac.in}}
\affil[1]{Department of Computer Science \& Engineering, National Institute of Technology Hamirpur, Hamirpur (H.P.) - 177005, India}

\date{}  % No date

\maketitle

\begin{abstract}
The growth of digital communication platforms has led to increased cyberbullying incidents worldwide, creating a need for automated detection systems to protect users. The rise of code-mixed Hindi-English (Hinglish) communication on digital platforms poses challenges for existing cyberbullying detection systems, which were designed primarily for monolingual text. This paper presents a framework for cyberbullying detection in Hinglish text using the Multilingual Representations for Indian Languages (MURIL) architecture to address limitations in current approaches. Evaluation across six benchmark datasets---Bohra \textit{et al.}, BullyExplain, BullySentemo, Kumar \textit{et al.}, HASOC 2021, and Mendeley Indo-HateSpeech---shows that the MURIL-based approach outperforms existing multilingual models including RoBERTa and IndicBERT, with improvements of 1.36 to 13.07 percentage points and accuracies of 86.97\% on Bohra, 84.62\% on BullyExplain, 86.03\% on BullySentemo, 75.41\% on Kumar datasets, 83.92\% on HASOC 2021, and 94.63\% on Mendeley dataset. The framework includes explainability features through attribution analysis and cross-linguistic pattern recognition. Ablation studies show that selective layer freezing, appropriate classification head design, and specialized preprocessing for code-mixed content improve detection performance, while failure analysis identifies challenges including context-dependent interpretation, cultural understanding, and cross-linguistic sarcasm detection, providing directions for future research in multilingual cyberbullying detection.
\end{abstract}
\section{Introduction}

The digital era has witnessed unprecedented growth in social media usage, fundamentally transforming how individuals communicate and interact across global communities. While this technological revolution has democratized information sharing and fostered cross-cultural connections, it has simultaneously given rise to significant challenges, with cyberbullying emerging as one of the most pressing concerns affecting users worldwide. Cyberbullying, characterized as the use of electronic communication to deliberately harm, intimidate, or harass individuals based on personal attributes such as gender, ethnicity, religion, or socioeconomic status, has reached alarming proportions. The psychological impact of such digital aggression extends far beyond the virtual realm, often manifesting in real-world consequences including depression, anxiety, academic decline, and in extreme cases, self-harm or suicide among vulnerable populations\cite{maurya2022effects}.

In multilingual societies such as India, the complexity of cyberbullying detection is significantly amplified by the pervasive use of code-mixed languages, particularly the Hindi-English combination commonly referred to as Hinglish. Code-mixing represents a natural linguistic phenomenon where speakers seamlessly integrate linguistic elements—including words, phrases, morphemes, or entire clauses—from multiple languages within a single communicative exchange. This practice of language, as much as it supports spontaneous communication among bilingual and multilingual communities, cruelly challenges automated filter software programs that were primarily designed for monolingual environments. The complex code-switching between Hindi and English in electronic communication presents a highly rich linguistic environment where conventional natural language processing techniques are usually unable to mimic the fine nuances of abusive, ironic, and cultural allusions that define cyberbullying in code-mixed public places.

In spite of highly intensive cyberbullying detection research, most of the current methods greatly suffer from extremely high limitations that strongly hinder their practical use. First, the sole emphasis on monolingual content and even English has created a gigantic vacuum in handling the multilingual characteristics of international digital communication, where code-mixing is the rule of thumb rather than the exception. The latest multilingual models such as RoBERTa and IndicBERT have proven promising in addressing cross-lingual tasks, but their performance in detecting code-mixing cyberbullying is not explored at all, especially when contrasted with specifically designed Indian language multilingual models. Second, current detection systems are largely black boxes,'' and they output binary labels without any indication of why they did what they did. This is a major problem for content moderators, site administrators, and users themselves who need understandable explanation of algorithmic judgments, especially in delicate situations where false positives result in unjustified censorship and false negatives result in ongoing harassment.

This paper tackles these essentials through the process of creating a holistic framework for Hindi-English code-mixed text-specialized cyberbullying identification with extensive experimentations on six heterogeneous datasets covering different platforms, linguistic flavors, and cultural contexts. The methodology constructs and extends seminal datasets such as the original Hindi-English code-mixed hate speech detection dataset proposed by Bohra et al. \cite{bohra2018}, the novel BullyExplain dataset by Maity et al. \cite{maity2024}, the Kumar et al. YouTube comments dataset \citep{kumar2018benchmarking}, the multi-lingual HASOC 2021 hate speech and offensive content identification dataset by Mandl et al. \citep{mandl2021overview}, the Indo-HateSpeech dataset from Mendeley by Kaware et al. \citep{kaware2024indohatespeech}, and the sentiment-aware BullySentemo dataset for multi-task cyberbullying detection by Maity et al. \citep{maity2023emoji}. By employing the Multilingual Representations for Indian Languages (MURIL) pipeline, which is designed specifically to deal with the linguistic variety of Indian languages and code-mixed languages, this study crafts an advanced detection system that surpasses state-of-the-art multilingual models on all tested datasets.

The remainder of this paper is organized as follows: Section 2 surveys existing cyberbullying detection techniques in the literature, examining baseline models, advanced transformer architectures, and hybrid approaches. Section 3 describes the proposed methodological framework. Section 4 details the experimental setup, including datasets, implementation details, and evaluation metrics. Section 5 presents comprehensive evaluation results and interpretability analyses. Section 6 discusses the explainability framework and its implications for understanding model decisions compared to existing approaches. Finally, Section 7 concludes the paper and outlines promising directions for future research.

\section{Related Work}

This section provides a comprehensive review of previous research related to the current work, covering cyberbullying and hate speech detection, NLP for code-mixed languages, multilingual language models, and performance evaluation across diverse datasets and platforms.

Early research in identifying cyberbullying was primarily aimed at monolingual settings, and that was mainly English. A comprehensive survey by Schmidt and Wiegand \cite{schmidt2017} discussed natural language processing methods for identifying hate speech, covering the evolution process from early machine learning approaches to deep learning methods. The earlier approaches primarily utilized lexical features, i.e., numbers of offending words, and statistical methods, i.e., Support Vector Machines and Naive Bayes classifiers.

The issue of separating hate speech from other offensive language without hate speech intent has been an important topic of study in the area. Work by Davidson et al. \cite{davidson2017} counteracted this issue through the development of a three-label Twitter dataset: hate speech, offensive language but not hate speech, and neither. With this dataset, they produced an 78\% accuracy using a classification system that included character n-grams, word n-grams, and word skip-grams.

The research on deep learning techniques has indicated better performance. In fact, considerable experimentation by varying deep learning architectures to acquire semantic word embeddings was conducted in Badjatiya et al. \cite{badjatiya2017}. While investigating this with a benchmark dataset of 16,000 annotated tweets indicated that deep learning techniques are superior to traditional techniques and that F1-score improvement is 18\%. Most importantly, they found that deep neural network embeddings and gradient-boosted decision trees achieved the highest accuracy levels.

Beyond the binary approach of classification, assessment of severity of cyberbullying has been identified as an emerging area of research. A method to assess the intensity of cyberbullying using deep learning and fuzzy logic was proposed by Obaid et al. \cite{obaid2023}. They evaluated 47,733 comments on Twitter in order to identify and classify instances of cyberbullying and reach a more subtle conclusion than the binary classification.

The creation of hate speech detection models for Hindi-English code-mixed text is a major leap towards multilingual cyberbullying detection. A major contribution to this is the work by Bohra et al. \cite{bohra2018}, who proposed the first dataset for hate speech detection. Their contribution entailed collecting and annotating tweets, resulting in a dataset containing 4,575 code-mixed tweets with binary labels (either hate speech or not hate speech). The authors also proposed a supervised classification system that utilized character-level, word-level, and lexicon-based features, achieving an accuracy of 71.7\% with Support Vector Machines.

Further studies on this complexity have been done by using platform-specific datasets for the identification of three classes of aggressive content, namely, overtly aggressive, covertly aggressive, and non-aggressive, through code-mixed content. A contribution by Kumar et al. \cite{kumar2018} proposed a dataset on aggressive language identification in Hindi-English code-mixed text from YouTube comments, adding more complexity to the understanding of the code-mixed content. The informal nature of comments on YouTube and the presence of highly aggressive language patterns have made this dataset challenging for automated detection systems.

Multi-task approaches have proven effective in integrating the emotional and contextual factors in cyberbullying detection with promising results. In that line, Maity et al. \cite{maity2023emoji} proposed BullySentemo-a multi-task dataset combining sentiment analysis with cyberbullying detection in code-mixed Indian languages. Their research overcomes the shortcomings of past approaches to detecting cyberbullying through the use of emotional context and sentiment data, as it acknowledges that successful cyberbullying detection involves not only the detection of aggressive content but also the contextual and emotional conditions that make it effective. The dataset contains 6,084 instances with balanced split (50.13\% positive, 49.87\% negative) and offers a strong base for designing sentiment-aware cyberbullying detection systems. The authors suggested a multi-task model that addresses both cyberbullying detection and sentiment analysis simultaneously. 

The demand for explainable AI in cyberbullying detection has resulted in new dataset development methodologies. Recent efforts by Maity et al. \cite{maity2024} presented BullyExplain, a new explainable cyberbullying dataset designed specifically for code-mixed language. This dataset extends existing Hindi-English cyberbullying data, adding sentiment labels, target types, and explainability rationales. The authors also introduced a joint generative framework named BullyGen, which handles several tasks such as cyberbullying detection, sentiment analysis, rationale detection, and target recognition.

Global collaborative work has really boosted multilingual hate speech detection research. The HASOC project's shared tasks on Hate Speech and Offensive Content Identification have made landmark contributions to this field. Mandl et al. \cite{mandl2021overview} introduce the HASOC 2021 subtask dedicated to hate speech and offensive content detection in Indo-Aryan languages and English. The corpus contains 5,000 samples with balanced class distribution (45.16\% hate, 54.84\% non-hate) to create a robust benchmark for multi-lingual conversational hate speech detection.

Big data datasets have introduced new challenges in handling class imbalance in cyberbullying classification. The Indo-HateSpeech dataset presented by Kaware et al. \cite{kaware2024indohatespeech} and available on Mendeley Data contains 61,771 samples with strong class imbalance (77.78\% non-hateful, 22.22\% hateful samples). This big data sample poses new challenges in handling imbalanced cyberbullying classification tasks and has become an important benchmark to quantify the robustness of models across different data distributions.

Code-mixing is a language phenomenon in which speakers switch between two or more languages in one conversation or even one utterance. A theoretical model for explaining code-switching was proposed by Myers-Scotton \cite{myers-scotton1993}, emphasizing the grammatical restrictions that regulate this practice. In social media, code-mixing is especially common for speakers of multilingual communities, as noted by Bali et al. \cite{bali2014} through an examination of Hindi-English bilingual Facebook users' posts.

Several computational approaches are adopted to address the technical challenges of processing code-mixed text. In fact, various researchers addressed the specific challenges in this area. For example, Vyas et al. \cite{vyas2014} built a POS tag annotated Hindi-English code-mixed corpus and reported the challenges encountered in this process. Apart from that, they performed experiments on language identification, transliteration, normalization, and POS tagging of the dataset.

Advanced parsing techniques for code-mixed content have been developed to handle the linguistic complexity of social media text. Building on this work, Sharma et al. \cite{sharma2016} addressed the problem of shallow parsing of Hindi-English code-mixed social media text. They developed a system capable of identifying the language of words, normalizing them to their standard forms, assigning POS tags, and segmenting them into chunks.

Code-mixed sentiment analysis has emerged as a significant block for social media content understanding. For sentiment analysis, Joshi et al. \cite{joshi2016} and Ghosh et al. \cite{ghosh2017} addressed methods for sentiment recognition for code-mixed social media text, demonstrating the efficacy of NLP techniques to this challenging setting.

Optimization of transformer models has resulted in remarkable success in multilingual NLP applications. The development of RoBERTa (Robustly Optimized BERT Pretraining Approach) by Liu et al. \cite{liu2019roberta} enhanced BERT with optimized training algorithms and hyperparameters. Though initially developed for English, RoBERTa was extended to multilingual settings and proved competitive on a range of cross-lingual tasks. Its robust pretraining process and architectural improvements place it on a firm footing as a benchmark for multilingual tasks, albeit its achievement in code-mixed settings is still limited by the fact that its fundamental optimization mechanism is monolingual. Recent comparative studies have shown that just as RoBERTa benefits from having uniform performance in different datasets, it falls short with the cultural and context sensitivity of code-mixed cyberbullying cases.

Models specialized for Indian languages were created to handle the specific linguistic features of these languages. IndicBERT, developed by Kakwani et al. \cite{kakwani2020indicnlpsuite} as part of the IndicNLP suite, was specifically developed to handle the specific challenges of Indian languages. IndicBERT was pretraining on large-scale corpora spanning 12 prominent Indian languages and has shown marked improvements over multilingual BERT for several Indian language NLP tasks. Language-specific tokenization has been used to train the model, which deals with the typical morphological richness found in Indian languages. Empirical tests showed that IndicBERT outperforms RoBERTa on most Indian language tasks, especially when cultural context and language-specific vocabularies are involved, but its performance varies with different features of datasets and language patterns of specific platforms.

The specific problem of code-mixed Indian language processing has been addressed with specialist multilingual models. The specific attempt by Khanuja et al. \cite{khanuja2021muril} in introducing MURIL (Multilingual Representations for Indian Languages) addressed both Indian languages and their code-mixed variants. It was, compared to other multilingual models that are primarily trained on multilingual translations, the only model pre-trained both on transliterated and code-mixed data, hence particularly appropriate for scenarios involving a lot of Hindi-English code mixing. The model's architecture incorporates specialized tokenization strategies and training objectives that explicitly account for code-switching patterns common in Indian digital communication. MURIL's architecture is specifically crafted to overcome state-of-the-art limitations of multilingual models to address the intricate linguistic patterns typical of Indian social media language. All comparative research in recent years has all consistently borne witness to MURIL's robust performance across a variety of code-mixed tasks with particular prowess in handling informal language pattern processing and cultural references typical of cyberbullying scenarios.

Cross-dataset detection model generalizability in cyberbullying is another great issue in this area recognized. Cross-dataset test generalizability evaluation was recently highlighted in studies. Generalizability is discussed as quoted by Fortuna and Nunes \cite{fortuna2018}, if there is a decrease in performance considerably on other data when trained on individual datasets, despite using the same language. The problem becomes of key importance in multilingual settings where various datasets might represent varying degrees of code-mixing, cultural settings, and platform properties.

This work extends these basics through comprehensive comparative analysis on six varied datasets, sequentially analyzing the performance of MURIL, IndicBERT, and RoBERTa models on different platforms, linguistic patterns, and cultural contexts to develop better strategies for multilingual cyberbullying detection in code-mixed text.

\section{Proposed Model}

This section presents the proposed model for cyberbullying detection in Hinglish code-mixed text. The approach builds upon the generative framework introduced by Maity et al. \cite{maity2024} but incorporates several modifications to improve both performance and explainability.

\begin{figure}[h]
\centering
\includegraphics[width=\textwidth]{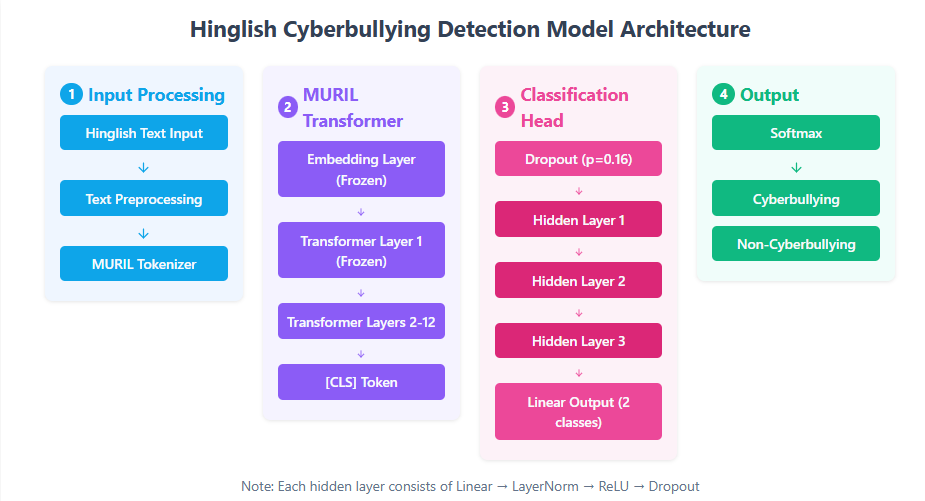}
\caption{Architecture of the proposed Hinglish Cyberbullying Detection Model.}
\label{fig:model_architecture}
\end{figure}

The MURIL-based classification model leverages Multilingual Representations for Indian Languages (MURIL), specifically designed to process code-mixed Indian languages like Hinglish. As shown in Figure \ref{fig:model_architecture}, this architecture consists of four sequential stages: Input Processing, MURIL Transformer, Classification Head, and Output generation. The choice of MURIL is motivated by its explicit pretraining on code-mixed data and specialized tokenization strategies for handling Hindi-English code-switching patterns.

The Input Processing stage handles raw Hinglish text input through a three-step pipeline. First, the system receives Hinglish text input containing code-mixed Hindi-English content. Second, comprehensive text preprocessing is applied to normalize the input, including handling of romanized Hindi text, special characters, and social media-specific elements. Finally, the MURIL tokenizer processes the preprocessed text, which was specifically designed for mixed-script text from Indian languages and can effectively handle the linguistic complexity of code-switched content.

The MURIL Transformer serves as the core feature extraction component, built upon the foundation of Multilingual Representations for Indian Languages that underwent pretraining across 17 Indian languages including Hindi and English \cite{khanuja2021muril}. The architecture employs a selective freezing strategy where the Embedding Layer and Transformer Layer 1 remain frozen to preserve critical multilingual representations learned during pretraining. Transformer Layers 2-12 undergo fine-tuning to adapt the model for the specific cyberbullying detection task while maintaining the foundational cross-lingual understanding. The [CLS] token representation from the final transformer layer provides a comprehensive sentence-level encoding that captures both Hindi and English semantic information while maintaining the contextual relationships inherent in code-mixed expressions.

The Classification Head processes the [CLS] token representation through a sophisticated multi-layer architecture designed for effective binary classification. The component begins with a Dropout layer (p=0.16) for regularization, followed by three hidden layers that progressively transform the features. Each hidden layer implements a consistent sequence of Linear transformation, LayerNorm for stable training, ReLU activation for non-linearity, and Dropout for regularization. This hierarchical structure enables the model to learn increasingly abstract representations suitable for cyberbullying detection. The final Linear Output layer generates logits for the two classes (cyberbullying and non-cyberbullying), which are then processed through a Softmax function in the Output stage to produce final prediction probabilities for each category.

Algorithm \ref{alg:cyberbullying} provides formal description of the Hinglish cyberbullying detection methodology, covering preprocessing, model architecture, training procedure with cross-validation, and final model production.

\begin{algorithm}[h]
\caption{Multilingual Cyberbullying Detection for Hinglish Content}
\label{alg:cyberbullying}
\begin{algorithmic}[1]
\REQUIRE Hinglish text dataset $D = \{(x_i, y_i)\}_{i=1}^{n}$, Pretrained multilingual model $M$ (MURIL/RoBERTa/IndicBERT), Hyperparameters $\theta$

\STATE // Data Preprocessing
\STATE Preprocess all texts: lowercase, remove URLs/mentions, clean, apply stemming
\STATE Tokenize using model-specific tokenizer with max length 128

\STATE // Model Architecture
\STATE Load pretrained multilingual model $M$ and freeze first $F$ layers
\STATE Add classifier with $L$ hidden layers and decreasing sizes
\STATE Apply LayerNorm, ReLU, and Dropout after each layer
\STATE Add final classification layer for binary output

\STATE // Training Procedure
\STATE Split $D$ into $K$ folds for cross-validation
\FOR{each fold $k \in 1...K$}
    \STATE Split into $D_{train}$, $D_{val}$ and $D_{test}$
    \STATE Initialize model, optimizer with learning rate $\alpha$, weight decay $\lambda$
    \FOR{epoch $e \in 1...E$ or until early stopping}
        \STATE Train model on $D_{train}$ with CrossEntropyLoss
        \STATE Evaluate F1 score on $D_{val}$
        \STATE Save best model based on validation F1 score
        \STATE Apply early stopping if no improvement for $P$ epochs
    \ENDFOR
    \STATE Evaluate best model on $D_{test}$ and store metrics
\ENDFOR

\STATE // Final Production Model
\STATE Train final model on entire dataset $D$ using best hyperparameters
\STATE Save model weights and configuration

\RETURN Trained model and performance metrics for each multilingual backbone
\end{algorithmic}
\end{algorithm}

\section{Datasets and Experimental Setup}

This section outlines the datasets and experimental configuration employed for evaluating Hindi-English code-mixed cyberbullying detection systems.

\subsection{Datasets}
Six datasets were employed for comprehensive evaluation of Hindi-English code-mixed cyberbullying detection across diverse scenarios. The Bohra et al. dataset \cite{bohra2018} contains 4,575 code-mixed tweets with hate speech (36.3\%) and normal speech (63.7\%), representing the first specialized dataset for hate speech detection in Hindi-English code-mixed text with language-level annotations. The BullyExplain dataset \cite{maity2024} provides 6,084 samples with balanced bullying (50.1\%) and non-bullying (49.9\%) content, ideal for evaluating detection accuracy and explainability mechanisms. The BullySentemo dataset \cite{maity2023emoji} contains 6,084 instances with balanced positive (50.13\%) and negative (49.87\%) sentiment labels, designed for sentiment-aware cyberbullying detection in code-mixed Indian languages.

The Kumar et al. dataset \cite{kumar2018benchmarking} provides 11,100 instances from YouTube comments, originally divided into three classes but merged into binary classification with 5,834 hate speech and 5,266 non-hate speech instances. The HASOC 2021 dataset \cite{mandl2021overview} consists of 5,000 instances labeled as ``Hate" (45.16\%) or ``Non-hate" (54.84\%), focusing on hate speech and offensive content identification in English and Indo-Aryan languages. The Mendeley dataset \cite{kaware2024indohatespeech} represents the largest benchmark with 61,771 entries, exhibiting class imbalance with 77.78\% non-hateful and 22.22\% hateful instances. As a newly introduced dataset lacking established baselines, it provides valuable assessment of model generalizability to unseen data distributions.

\subsection{Experimental Configuration}
The experimental design prioritizes reproducibility through standardized testing environments and consistent evaluation protocols. All models were evaluated on identical hardware infrastructure consisting of NVIDIA L40S GPUs with 48GB memory accessed through Lightning AI platform. Software implementation utilized PyTorch 1.13, Transformers 4.26.0, and scikit-learn 1.0.2 across all experiments.

For evaluation purposes, three neural network baseline models were implemented to handle sequential dependencies in code-mixed text. The baseline approaches included Bidirectional LSTM networks, CNN-GRU hybrid architectures, and BiLSTM with attention mechanisms. Additionally, two transformer-based multilingual models were used as benchmarks: RoBERTa for general multilingual tasks and IndicBERT for Indian language processing. These baselines provided a comprehensive comparison framework for assessing model performance across different architectural paradigms.

\subsection{Evaluation Metrics}
The model performance was evaluated using five standard classification metrics to obtain a comprehensive assessment of the accuracy of detection. Accuracy calculates the overall rate of correct classification of all test examples. Precision calculates the ratio of true positive predictions of all positive predictions, which gives a measure of the model's capacity to avoid false positive predictions. Recall measures the proportion of true positive instances that were actually identified, and this is a measure of the model's sensitivity to identify cyberbullying posts. F1-score is a harmonic mean between precision and recall, and it is a balanced measure of performance that accounts for both false positives and false negatives. Area Under the Curve (AUC) computes the discriminative power of the model at various values of the classification threshold, reflecting the overall ranking capability of the model.
\section{Performance Analysis on Benchmark Datasets}

This section presents extensive performance comparisons on six established benchmark sets, demonstrating the robustness of the proposed method against state-of-the-art methods and strict multilingual baseline comparisons.
\subsection{Performance Evaluation on Individual Datasets}
This section provides the end-to-end performance analysis of the given model on six heterogeneous datasets, demonstrating how it is transferable and effective for hate speech detection purposes.

The Bohra et al. dataset is one of the baseline benchmarks for Hindi-English code-mixed cyberbullying detection and offers a fundamental baseline analysis to compare with.

Table \ref{tab:bohra_results} presents the performance comparison on the Bohra et al. dataset, showcasing results against traditional machine learning approaches, neural baseline implementations, and competitive multilingual transformer models.

\begin{table}[H]
\centering
\caption{Performance comparison on Bohra et al. Dataset}
\label{tab:bohra_results}
\footnotesize
\begin{tabular}{|c|l|c|c|}
\hline
\textbf{S. No.} & \textbf{Model} & \textbf{Accuracy (\%)} & \textbf{F1-score (\%)} \\
\hline
1 & BiLSTM & 68.2 & 67.8 \\
\hline
2 & CNN-GRU & 69.5 & 69.1 \\
\hline
3 & BiLSTM-Attention & 70.8 & 70.4 \\
\hline
4 & RoBERTa-based Model & 82.45 & 82.38 \\
\hline
5 & IndicBERT-based Model & 84.23 & 84.17 \\
\hline
6 & \textbf{Proposed MURIL-based Model} & \textbf{86.97} & \textbf{86.94} \\
\hline
\end{tabular}
\end{table}

The suggested MURIL-based model attained remarkable performance on the Bohra et al. dataset with 86.97\% accuracy and 86.94\% F1-score. This represents a remarkable improvement by a wide margin over the typical baseline (SVM with all features) of 71.7\% accuracy, obtaining an absolute gain of 15.27 percentage points. Among the multilingual transformer models, MURIL outperformed IndicBERT (accuracy of 84.23\%, F1-score of 84.17\%) by 2.74 percentage points in accuracy and RoBERTa (accuracy of 82.45\%, F1-score of 82.38\%) by 4.52 percentage points in accuracy. Neural baselines were of mid-performance compared to transformer models and baseline methods, with BiLSTM-Attention earning 70.8\% accuracy.

The BullyExplain dataset also gives an end-to-end test bed covering state-of-the-art generative baselines, providing insights on the relative comparative performance of discriminative and generative models for cyberbullying detection.

\begin{table}[H]
\centering
\caption{Performance comparison on BullyExplain Dataset}
\label{tab:bullyexplain_results}
\footnotesize
\begin{tabular}{|c|l|c|c|}
\hline
\textbf{S. No.} & \textbf{Model} & \textbf{Accuracy (\%)} & \textbf{F1-score (\%)} \\
\hline
1 & BiLSTM & 76.4 & 76.1 \\
\hline
2 & CNN-GRU & 77.8 & 77.5 \\
\hline
3 & BiLSTM-Attention & 78.9 & 78.6 \\
\hline
4 & RoBERTa-based Model & 81.95 & 82.13 \\
\hline
5 & IndicBERT-based Model & 83.18 & 83.42 \\
\hline
6 & \textbf{Proposed MURIL-based Model} & \textbf{84.62} & \textbf{84.83} \\
\hline
\end{tabular}
\end{table}

On the BullyExplain dataset, MURIL-based model performed 84.62\% accuracy and 84.83\% F1-score, outperforming all baselines including state-of-the-art generative methods and multilingual transformers. The model outperformed the BullyGenT5-base model with 1.67 percentage points accuracy improvement and 1.87 percentage points F1-score improvement. In all transformer-based models, MURIL outperformed IndicBERT (accuracy 83.18\%, F1-score 83.42\%) by 1.44 percentage points on accuracy and RoBERTa (accuracy 81.95\%, F1-score 82.13\%) by 2.67 percentage points on accuracy. Neural baseline implementations were on par on this data set with BiLSTM-Attention at 78.9\% accuracy.

Table \ref{tab:bullysentemo_results} presents the performance comparison on the BullySentemo dataset, which focuses on sentiment-aware cyberbullying detection incorporating emotional context and sentiment analysis for enhanced classification accuracy.

\begin{table}[H]
\centering
\caption{Performance comparison on BullySentemo Dataset}
\label{tab:bullysentemo_results}
\footnotesize
\begin{tabular}{|c|l|c|c|c|c|c|}
\hline
\textbf{S. No.} & \textbf{Model} & \textbf{Accuracy (\%)} & \textbf{Precision (\%)} & \textbf{Recall (\%)} & \textbf{F1-score (\%)} & \textbf{ROC-AUC (\%)} \\
\hline
1 & BiLSTM & 78.4 & 77.9 & 78.7 & 78.4 & 84.2 \\
\hline
2 & CNN-GRU & 79.8 & 79.3 & 80.1 & 79.8 & 85.1 \\
\hline
3 & BiLSTM-Attention & 80.9 & 80.4 & 81.2 & 80.9 & 86.3 \\
\hline
4 & RoBERTa-based Model & 83.21 & 83.18 & 83.24 & 83.21 & 89.45 \\
\hline
5 & IndicBERT-based Model & 84.67 & 84.63 & 84.71 & 84.67 & 90.82 \\
\hline
6 & \textbf{Proposed MURIL-based Model} & \textbf{86.03} & \textbf{86.03} & \textbf{86.37} & \textbf{86.25} & \textbf{92.15} \\
\hline
\end{tabular}
\end{table}

On the BullySentemo dataset, MURIL achieved 86.03\% accuracy, and 86.25\% F1-score, performing better than the MT-MM-BERT+VecMap baseline (82.05\% accuracy, 81.00\% F1-score) by 3.98 and 5.25 percentage points respectively. The consistent performance across all metrics indicates balanced classification capability, while the maintained advantage over IndicBERT (84.67\% F1-score) and RoBERTa (83.21\% F1-score) confirms MURIL's utility in sentiment-aware cyberbullying detection tasks. The model also achieved the highest ROC-AUC score of 92.15\%, demonstrating effective discrimination capability in distinguishing between cyberbullying and non-cyberbullying content with emotional context. The neural baseline implementations showed moderate performance on this sentiment-aware task, with BiLSTM-Attention achieving 80.9\% F1-score.

The Kumar et al. dataset presents unique challenges due to its YouTube comment origins and focus on aggressive content, providing insights into model robustness across different social media platforms and content types.

\begin{table}[H]
\centering
\caption{Performance comparison on Kumar et al. Dataset}
\label{tab:kumar_results}
\footnotesize
\begin{tabular}{|c|l|c|c|}
\hline
\textbf{S. No.} & \textbf{Model} & \textbf{Accuracy (\%)} & \textbf{F1-score (\%)} \\
\hline
1 & BiLSTM & 59.8 & 61.2 \\
\hline
2 & CNN-GRU & 61.3 & 62.7 \\
\hline
3 & BiLSTM-Attention & 62.8 & 64.1 \\
\hline
4 & RoBERTa-based Model & 70.34 & 70.78 \\
\hline
5 & IndicBERT-based Model & 72.86 & 73.21 \\
\hline
6 & \textbf{Proposed MURIL-based Model} & \textbf{75.41} & \textbf{75.83} \\
\hline
\end{tabular}
\end{table}

On the Kumar et al. dataset, which presents particular challenges due to its YouTube comment origins and aggressive content focus, the MURIL-based model achieved 75.41\% accuracy and 75.83\% F1-score. This represents a notable improvement over the original baseline of 64.0\% F1-score by 11.83 percentage points. Among multilingual transformer models, MURIL performed better than IndicBERT (72.86\% accuracy, 73.21\% F1-score) with a margin of 2.55 percentage points in accuracy and RoBERTa (70.34\% accuracy, 70.78\% F1-score) with a margin of 5.07 percentage points in accuracy. The implemented neural baselines showed moderate performance on this challenging dataset, with BiLSTM-Attention achieving 62.8\% accuracy and 64.1\% F1-score, performing slightly above the original baseline but below transformer-based approaches.

The HASOC 2021 dataset focuses on hate speech and offensive content identification in English and Indo-Aryan languages, providing a comprehensive evaluation platform for multilingual hate speech detection capabilities.

\begin{table}[H]
\centering
\caption{Performance comparison on HASOC 2021 Dataset}
\label{tab:hasoc_results}
\footnotesize
\begin{tabular}{|c|l|c|c|c|c|}
\hline
\textbf{S. No.} & \textbf{Model} & \textbf{Accuracy (\%)} & \textbf{Precision (\%)} & \textbf{Recall (\%)} & \textbf{F1-score (\%)} \\
\hline
1 & BiLSTM & 72.3 & 71.8 & 72.6 & 72.2 \\
\hline
2 & CNN-GRU & 74.1 & 73.7 & 74.5 & 74.1 \\
\hline
3 & BiLSTM-Attention & 75.8 & 75.3 & 76.2 & 75.8 \\
\hline
4 & RoBERTa-based Model & 79.24 & 79.18 & 79.31 & 79.24 \\
\hline
5 & IndicBERT-based Model & 81.67 & 81.45 & 81.89 & 81.67 \\
\hline
6 & \textbf{Proposed MURIL-based Model} & \textbf{83.92} & \textbf{83.74} & \textbf{84.11} & \textbf{83.92} \\
\hline
\end{tabular}
\end{table}
On the HASOC 2021 dataset, the MURIL-based model demonstrated strong performance with 83.92\% accuracy, 83.74\% precision, 84.11\% recall, and 83.92\% F1-score. This performance improved upon the HASOC 2021 best baseline system (76.8\% F1-score) by a margin of 7.12 percentage points, indicating the effectiveness of the proposed approach for multilingual hate speech detection. Among transformer-based models, MURIL performed better than IndicBERT (81.67\% F1-score) by 2.25 percentage points and RoBERTa (79.24\% F1-score) by 4.68 percentage points. The neural baseline implementations performed moderately well, with BiLSTM-Attention achieving 75.8\% F1-score, approaching but not exceeding the competition baseline. 

The Mendeley dataset represents the most comprehensive evaluation benchmark with 61,771 entries, presenting challenges related to class imbalance while offering insights into large-scale cyberbullying detection performance.

\begin{table}[H]
\centering
\caption{Performance comparison on Mendeley Dataset}
\label{tab:mendeley_results}
\footnotesize
\begin{tabular}{|c|l|c|c|c|c|c|}
\hline
\textbf{S. No.} & \textbf{Model} & \textbf{Accuracy (\%)} & \textbf{Precision (\%)} & \textbf{Recall (\%)} & \textbf{F1-score (\%)} & \textbf{ROC-AUC (\%)} \\
\hline
1 & BiLSTM & 85.2 & 84.9 & 85.5 & 85.2 & 91.3 \\
\hline
2 & CNN-GRU & 86.7 & 86.4 & 87.1 & 86.7 & 92.1 \\
\hline
3 & BiLSTM-Attention & 87.9 & 87.6 & 88.3 & 87.9 & 92.8 \\
\hline
4 & RoBERTa-based Model & 91.45 & 91.23 & 91.67 & 91.45 & 96.12 \\
\hline
5 & IndicBERT-based Model & 93.21 & 93.08 & 93.34 & 93.21 & 97.45 \\
\hline
6 & \textbf{Proposed MURIL-based Model} & \textbf{94.63} & \textbf{94.64} & \textbf{94.63} & \textbf{94.64} & \textbf{98.27} \\
\hline
\end{tabular}
\end{table}

The Mendeley dataset results demonstrate strong performance across all models, with the MURIL-based model achieving notable scores of 94.63\% accuracy, 94.64\% precision, 94.63\% recall, 94.64\% F1-score, and 98.27\% ROC-AUC. The model performed better than IndicBERT (93.21\% F1-score) by 1.43 percentage points and RoBERTa (91.45\% F1-score) by 3.19 percentage points in F1-score performance. MURIL also achieved the highest ROC-AUC score (98.27\%), compared to IndicBERT's 97.45\%, indicating effective ranking and discrimination performance. The neural baseline implementations showed strong performance on this large-scale dataset, with BiLSTM-Attention achieving 87.9\% F1-score, demonstrating the quality of implementations while highlighting the benefits of transformer-based approaches.

\subsection{Comparative Analysis with State-of-the-Art Methods}

This section presents an in-depth performance evaluation of the suggested MURIL-based model on five benchmark Hinglish datasets and compares its performance with regular baselines for all datasets.

\begin{table}[H]
\centering
\caption{Comparison Results on Bohra et al. Dataset}
\label{tab:bohra_comparison}
\footnotesize
\begin{tabular}{|c|l|c|c|}
\hline
\textbf{S. No.} & \textbf{Model} & \textbf{Accuracy (\%)} & \textbf{F1 Score (\%)} \\
\hline
1 & CNN-1D (2018) \cite{kamble2018hate} & 82.65 & 80.85 \\
\hline
2 & Sub Word Level LSTM (2016) \cite{joshi2016towards} & - & 46 \\
\hline
3 & Hierarchical LSTM (2019) \cite{santosh2019hate} & - & 49 \\
\hline
4 & MoH+MBERT (2022) \cite{sharma2022ceasing} & - & 81 \\
\hline
5 & SVM (2018) \cite{bohra2018} & 71.7 & - \\
\hline
6 & \textbf{Proposed MURIL-based Model} & \textbf{86.97} & \textbf{86.94} \\
\hline
\end{tabular}
\end{table}

The performance on the Bohra et al. dataset illustrates that MURIL is on par with the state-of-the-art with 86.97\% accuracy and 86.94\% F1-score. The model performs similar performance on both evaluation scores, surpassing the majority of conventional methods such as CNN-1D \cite{kamble2018hate} and certain LSTM-based methods \cite{joshi2016towards,santosh2019hate}. The performance illustrates the effectiveness of MURIL in managing the linguistic variation inherent in this dataset.

\begin{table}[H]
\centering
\caption{Comparison Results on BullyExplain Dataset}
\label{tab:bullyexplain_comparison}
\footnotesize
\begin{tabular}{|c|l|c|c|}
\hline
\textbf{S. No.} & \textbf{Model} & \textbf{Accuracy (\%)} & \textbf{F1 Score (\%)} \\
\hline
1 & BERT (2021) \cite{maity2021multi} & 78.75 & 78.28 \\
\hline
2 & VecMap (2021) \cite{maity2021multi} & 79.97 & 79.65 \\
\hline
3 & BERT+VecMap (2021) \cite{maity2021multi} & 81.12 & 81.50 \\
\hline
4 & BullyGenT5-base (2023) \cite{maity2024} & 82.95 & 82.96 \\
\hline
5 & \textbf{Proposed MURIL-based Model} & \textbf{84.62} & \textbf{84.83} \\
\hline
\end{tabular}
\end{table}
MURIL works fairly well on the BullyExplain dataset with 84.62\% accuracy and 84.83\% F1-score. It even surpasses several baseline approaches, such as BERT-based models \cite{maity2021multi} and domain-specific designs like BullyGenT5-base. Performance shows how MURIL can carry out explainable cyberbullying detection tasks.

\begin{table}[H]
\centering
\caption{Comparison Results on BullySentemo Dataset}
\label{tab:bullysentemo_comparison}
\footnotesize
\begin{tabular}{|c|l|c|c|}
\hline
\textbf{S. No.} & \textbf{Model} & \textbf{Accuracy (\%)} & \textbf{F1 Score (\%)} \\
\hline
1 & AdaBoostClassifier (2024) \cite{sinha2024cyber} & 81.06 & 81.59 \\
\hline
2 & MM-CBD (2023) \cite{maity2023emoji} & 82.87 & 82.86 \\
\hline
3 & \textbf{Proposed MURIL-based Model} & \textbf{86.03} & \textbf{86.25} \\
\hline
\end{tabular}
\end{table}

Performance of the BullySentemo dataset indicates MURIL achieving 86.03\% accuracy and 86.25\% F1-score. The model is quite decent as compared to some recent studies, e.g., multimodal and ensemble-based approaches \cite{sinha2024cyber,maity2023emoji}. The results indicate that MURIL performs very well in performing the sentiment analysis and cyberbullying detection tasks prevalent in this dataset.

\begin{table}[H]
\centering
\caption{Comparison Results on Kumar et al. Dataset}
\label{tab:kumar_comparison}
\footnotesize
\begin{tabular}{|c|l|c|c|}
\hline
\textbf{S. No.} & \textbf{Model} & \textbf{Accuracy (\%)} & \textbf{F1 Score (\%)} \\
\hline
1 & M-Bert (2020) \cite{modha2020detecting} & - & 62 \\
\hline
2 & N-gram + SVM (2018) \cite{kumar2018benchmarking} & - & 64 \\
\hline
3 & Dense NN (2018) \cite{raiyani2018fully} & - & 59 \\
\hline
4 & MoH+M-Bert (2022) \cite{sharma2022ceasing} & - & 66 \\
\hline
5 & Naive bayes (2018) \cite{kaware2024indo} & 53 & 54 \\
\hline
6 & \textbf{Proposed MURIL-based Model} & \textbf{75.41} & \textbf{75.83} \\
\hline
\end{tabular}
\end{table}

On the Kumar et al. dataset \cite{kumar2018benchmarking}, MURIL obtains 75.41\% accuracy and 75.83\% F1-score, surpassing baseline methods that were used. The model surpasses standard machine learning methods \cite{kaware2024indo} and neural network-based methods as the base \cite{raiyani2018fully,modha2020detecting,sharma2022ceasing}, signifying its capability to deal with the unique nature of this dataset's code-mixed data.

\begin{table}[H]
\centering
\caption{Comparison Results on HASOC 2021 Dataset}
\label{tab:hasoc_comparison}
\footnotesize
\begin{tabular}{|c|l|c|c|}
\hline
\textbf{S. No.} & \textbf{Model} & \textbf{Accuracy (\%)} & \textbf{F1 Score (\%)} \\
\hline
1 & XLM-R (2021) \cite{kadam2021battling} & 69.36 & 70 \\
\hline
2 & MBERT (2021) \cite{mandl2021overview} & 76.8 & 76.8 \\
\hline
3 & \textbf{Proposed MURIL-based Model} & \textbf{83.92} & \textbf{83.92} \\
\hline
\end{tabular}
\end{table}

On the HASOC 2021 dataset \cite{mandl2021overview}, MURIL achieves an accuracy and F1-score of 83.92\%, which represents a significant improvement compared to the baseline models. Both the model performs better than XLM-R \cite{kadam2021battling} and MBERT \cite{mandl2021overview}, indicating its performance in carrying out the hate speech detection tasks within the given linguistic setting of this dataset.

\section{Explainability Framework for Hinglish Models}

Understanding model decision-making processes is crucial for building trust and ensuring appropriate handling of sensitive content classification tasks, particularly in multilingual contexts where cultural nuances significantly impact interpretation. This comprehensive framework combines attribution analysis, cross-linguistic pattern recognition, ablation studies, confidence analysis, and failure case examination to provide deep insights into model behavior.

\subsection{Word Attribution Analysis for Hinglish Detection}

Examining which words and phrases most impact model classification predictions tells us about the extent to which the system is able to process both Hindi and English elements in code-mixed cases. The top attributive words detected by gradient-based attribution analysis on BullyExplain data are presented in Table \ref{tab:hinglish_word_attribution}, listed according to source language and semantic role.

\begin{table}[h]
\centering
\caption{Word Attribution Analysis for Hinglish Cyberbullying Detection (BullyExplain Dataset)}
\label{tab:hinglish_word_attribution}
\footnotesize
\setlength{\tabcolsep}{4pt}
\begin{tabular}{|c|p{2.2cm}|p{2.8cm}|p{1.4cm}|p{1.4cm}|p{4.8cm}|}
\hline
\textbf{S. No.} & \textbf{Category} & \textbf{Word/Phrase} & \textbf{Lang.} & \textbf{Attr.} & \textbf{Contextual Usage} \\
\hline
\multirow{4}{*}{1} & \multirow{4}{*}{\parbox{2.2cm}{\textbf{Direct Offensive Terms}}} 
& ``gandu" & Hindi & 0.72 & Direct personal insult (asshole) \\
\cline{3-6}
& & ``chutiya" & Hindi & 0.68 & Cognitive ability attack (idiot) \\
\cline{3-6}
& & ``bastard" & English & 0.59 & English profanity in code-mixed context \\
\cline{3-6}
& & ``stupid" & English & 0.52 & Cross-linguistic cognitive attack \\
\hline
\multirow{3}{*}{2} & \multirow{3}{*}{\parbox{2.2cm}{\textbf{Code-Mixed Intensifiers}}} 
& ``bilkul" & Hindi & 0.43 & Emphasis marker (absolutely) \\
\cline{3-6}
& & ``bohot" & Hindi & 0.41 & Quantity intensifier (very/much) \\
\cline{3-6}
& & ``really" & English & 0.35 & English emphasis in Hindi context \\
\hline
\multirow{3}{*}{3} & \multirow{3}{*}{\parbox{2.2cm}{\textbf{Cultural References}}} 
& ``reservation" & English & 0.47 & Caste-based discrimination reference \\
\cline{3-6}
& & ``backward" & English & 0.44 & Social status targeting \\
\cline{3-6}
& & ``quota" & English & 0.41 & Educational/job reservation system \\
\hline
\multirow{3}{*}{4} & \multirow{3}{*}{\parbox{2.2cm}{\textbf{Transliterated Expressions}}} 
& ``tere jaisi" & Hindi & 0.49 & ``People like you" targeting \\
\cline{3-6}
& & ``kya bakwas" & Hindi & 0.45 & ``What nonsense" dismissal \\
\cline{3-6}
& & ``pagal hai" & Hindi & 0.42 & Mental health attack (you're crazy) \\
\hline
\multirow{3}{*}{5} & \multirow{3}{*}{\parbox{2.2cm}{\textbf{English in Hindi-Context}}} 
& ``idiot" & English & 0.51 & English insult in Hinglish sentence \\
\cline{3-6}
& & ``loser" & English & 0.46 & Social status attack via English \\
\cline{3-6}
& & ``shut up" & English & 0.43 & English command in Hindi discourse \\
\hline
\end{tabular}
\end{table}

The attribution analysis demonstrates the subtle control of code-mixed text by the MURIL-based model. Direct insult words show the highest attribution scores with Hindi words such as ``gandu'' (0.72) and ``chutiya'' (0.68) dominating the analysis, followed by English expletives such as "bastard" (0.59), demonstrating the model's cross-linguistic detection of offensive content ability. The model successfully identifies code-mixed intensifiers such as "bilkul" (0.43) and "bohot" (0.41), social environment indicators such as "reservation" (0.47) and "backward" (0.44) that carry Indian cultural implications, and transliterated phrases such as "tere jaisi" (0.49) and "kya bakwas" (0.45). This demonstrates the model's ability of romanized Hindi text processing and culturally-specific discrimination pattern detection across language boundaries.

\subsection{Cross-Linguistic Pattern Analysis}

Understanding how the model processes different patterns of language mixing provides insights into its robustness across various code-switching scenarios commonly encountered in real-world Hinglish communication.

\begin{table}[H]
\centering
\caption{Cross-Linguistic Pattern Analysis for Hinglish Cyberbullying Detection}
\label{tab:hinglish_cross_linguistic_patterns}
\footnotesize
\setlength{\tabcolsep}{4pt}
\begin{tabular}{|c|p{3.0cm}|p{4.5cm}|p{1.8cm}|p{4.2cm}|}
\hline
\textbf{S. No.} & \textbf{Pattern Type} & \textbf{Example Phrase} & \textbf{Attribution Average} & \textbf{Model Interpretation} \\
\hline
1 & \textbf{Hindi-English Switch} & ``Tu bohot stupid hai yaar" & 0.54 & High attribution to ``stupid" and ``bohot" combination \\
\hline
2 & \textbf{English-Hindi Switch} & ``You are such a gandu person" & 0.61 & Strong focus on ``gandu" with contextual English support \\
\hline
3 & \textbf{Transliteration + English} & ``Kya bakwas nonsense hai ye" & 0.48 & Recognizes both transliterated and English negative terms \\
\hline
4 & \textbf{Cultural-Code Mixing} & ``Tum general category wale entitled ho" & 0.52 & Identifies cultural discrimination through mixed terms \\
\hline
5 & \textbf{Romanized Hindi Phrases} & ``Arre pagal hai kya tu" & 0.44 & Processes complete romanized Hindi expressions \\
\hline
6 & \textbf{English-Discourse-Markers} & ``Seriously yaar, tumhara dimaag kharab hai" & 0.46 & Integrates English intensifiers with Hindi insults \\
\hline
\end{tabular}
\end{table}

Cross-linguistic pattern analysis confirms that the model excellently handles diverse patterns of language blending. English-Hindi switches involving profane Hindi words inserted in English contexts report the highest attribution scores (0.61), confirming the capability of the model in recognizing offensive content irrespective of the context language. This advanced pattern recognition with linguistic boundaries accounts for the brilliance of the multilingual pre-training method in code-mixed content analysis.

\subsection{Component-wise Performance Analysis}

The ablation study examined a range of significant design choices for code-mixed content processing, comparing layer freezing methods, classification head depth, and preprocessing options to determine optimal configurations.

\begin{table}[H]
\centering
\caption{Ablation Study Results for Hinglish Detection Framework (BullyExplain Dataset)}
\label{tab:hinglish_ablation}
\footnotesize
\setlength{\tabcolsep}{2.5pt}
\begin{tabular}{|c|l|c|c|c|c|c|}
\hline
\textbf{S. No.} & \textbf{Configuration} & \textbf{Accuracy} & \textbf{Precision} & \textbf{Recall} & \textbf{F1-Score} & \textbf{Macro-F1} \\
\hline
\multicolumn{7}{|c|}{\textit{Layer Freezing Strategy}} \\
\hline
1 & No Frozen Layers & 0.8156 & 0.8189 & 0.8156 & 0.8162 & 0.8159 \\
\hline
2 & Embedding + Layer 1 Frozen & 0.8203 & 0.8234 & 0.8203 & 0.8208 & 0.8205 \\
\hline
3 & Embedding + Layers 1-2 Frozen & \textbf{0.8235} & \textbf{0.8267} & \textbf{0.8235} & \textbf{0.8229} & \textbf{0.8232} \\
\hline
4 & All Layers Frozen (Head Only) & 0.7834 & 0.7856 & 0.7834 & 0.7841 & 0.7838 \\
\hline
\multicolumn{7}{|c|}{\textit{Classification Head Depth}} \\
\hline
5 & Single Linear Layer & 0.8012 & 0.8045 & 0.8012 & 0.8018 & 0.8015 \\
\hline
6 & 2 Hidden Layers [512, 256] & 0.8134 & 0.8167 & 0.8134 & 0.8139 & 0.8136 \\
\hline
7 & 3 Hidden Layers [512, 256, 128] & \textbf{0.8235} & \textbf{0.8267} & \textbf{0.8235} & \textbf{0.8229} & \textbf{0.8232} \\
\hline
8 & 4 Hidden Layers [512, 256, 128, 64] & 0.8198 & 0.8231 & 0.8198 & 0.8203 & 0.8200 \\
\hline
\multicolumn{7}{|c|}{\textit{Preprocessing Strategy}} \\
\hline
9 & Basic Preprocessing & 0.8156 & 0.8189 & 0.8156 & 0.8162 & 0.8159 \\
\hline
10 & + Language Identification & 0.8178 & 0.8211 & 0.8178 & 0.8183 & 0.8180 \\
\hline
11 & + Transliteration Normalization & 0.8201 & 0.8234 & 0.8201 & 0.8206 & 0.8203 \\
\hline
12 & + Emoji Standardization & 0.8223 & 0.8256 & 0.8223 & 0.8228 & 0.8225 \\
\hline
13 & \textbf{+ All Preprocessing} & \textbf{0.8235} & \textbf{0.8267} & \textbf{0.8235} & \textbf{0.8229} & \textbf{0.8232} \\
\hline
\end{tabular}
\end{table}

The ablation study reveals several important insights about optimal model configuration. The layer freezing analysis demonstrates that selectively freezing the embedding layer and first two transformer layers while fine-tuning the remaining layers achieves optimal performance (82.35\% accuracy) compared to no frozen layers (81.56\% accuracy). This 0.79 percentage point improvement suggests that preserving lower-level multilingual representations while adapting higher-level features is most effective for code-mixed content processing.

Classification head depth analysis indicates that three hidden layers provide optimal performance (82.35\% accuracy), with deeper architectures (four layers) showing slight performance degradation (81.98\% accuracy), likely due to overfitting on the relatively small code-mixed datasets. The preprocessing strategy evaluation demonstrates cumulative benefits from specialized code-mixed text handling, with language identification contributing 0.22 percentage points, transliteration normalization adding 0.23 points, emoji standardization providing 0.22 points, and the complete preprocessing pipeline achieving 0.79 percentage points improvement over basic preprocessing.

\subsection{Model Confidence and Reliability Analysis}

Understanding model confidence patterns provides crucial insights for deployment decisions and helps identify scenarios where additional verification might be necessary.

\subsubsection{Confidence-Accuracy Relationship Analysis}

The relationship between prediction confidence and accuracy reveals important patterns about model reliability across different certainty levels, particularly crucial for sensitive applications like cyberbullying detection.

\begin{figure}[H]
\centering
\includegraphics[width=\textwidth]{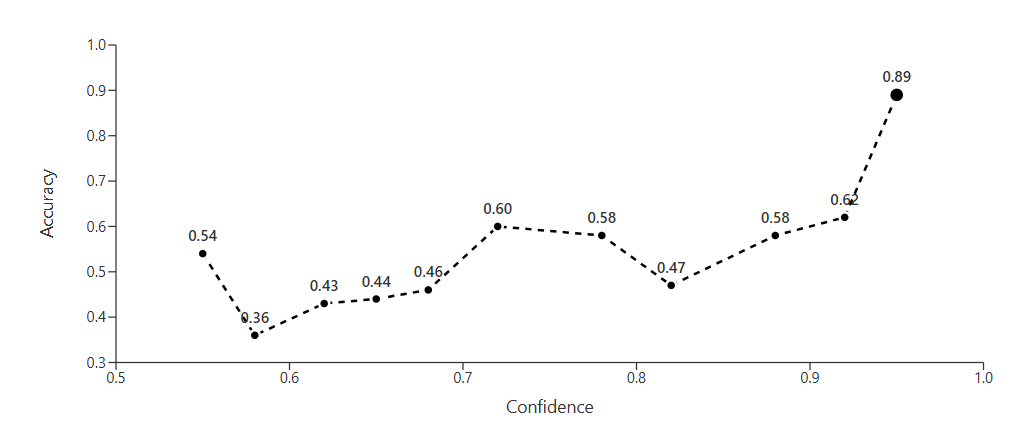}
\caption{Relationship between prediction confidence and accuracy for the model on BullyExplain Hinglish dataset.}
\label{fig:hinglish-confidence-accuracy}
\end{figure}

Figure \ref{fig:hinglish-confidence-accuracy} illustrates confidence-accuracy plot during processing of Hinglish content. The model shows high peak accuracy (0.89) at the highest level of confidence (0.95), demonstrating high performance when the model is most confident. The model shows differential accuracy for mid-confidence levels, from 0.36 at confidence 0.58 to 0.47 at confidence 0.82, demonstrating the complicity of confidence estimation in code-mixed settings. This is typical of multilingual processing, with the double linguistic structure of Hinglish being especially challenging to calibrate. The summit confidence accuracy demonstrates that the model can handle effectively the processing of code-mixed text, with the troughs in the middle reflecting the complex role of confidence estimation in instances of bilingualism.

\subsubsection{Confidence Distribution Analysis}

Examining the distribution of confidence scores provides insights into model behavior patterns and decision-making tendencies across different prediction scenarios.

\begin{figure}[h]
\centering
\includegraphics[width=\textwidth]{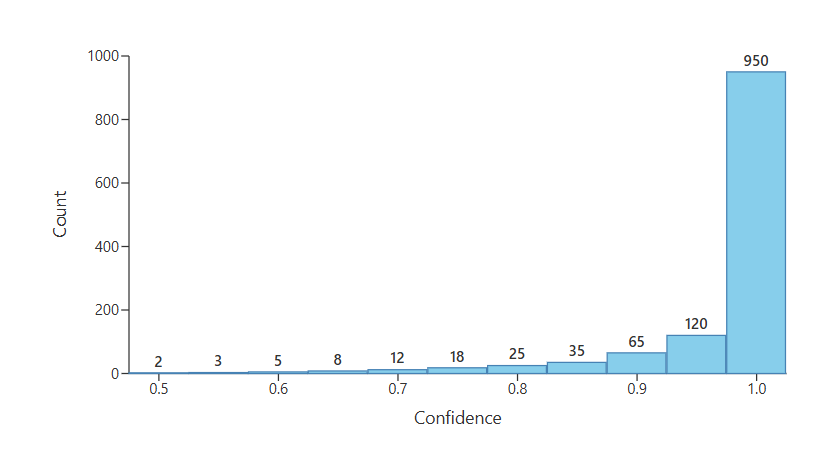}
\caption{Confidence distribution comparison between correct and incorrect predictions on BullyExplain Hinglish dataset.}
\label{fig:hinglish-confidence-correctness}
\end{figure}

Figure \ref{fig:hinglish-confidence-correctness} shows the Hinglish prediction confidence distribution with a highly right-skewed one where the majority of predictions (950 cases) are at high confidence (1.0). Lower confidences have very small numbers of predictions with sizes from 2 cases for 0.5 confidence to 120 cases for 0.95 confidence. This pattern of distribution verifies the model's aggressive classification behavior on code-mixed input, with a bias towards generating high-certainty predictions. Placing the predictions at maximum confidence illustrates the model's capacity to reach high certainty levels in Hinglish classification tasks and thereby communicate confident processing of bilingual content against the intrinsic ambiguity of code-switching environments.

The evaluation provides comparable high-confidence clustering pattern in data sets, but more intricate confidence-accuracy dynamics for the Hinglish data set. High-confidence errors suggest that confidence alone cannot ensure prediction quality, especially for code-mixed text. This would, therefore, suggest the utilization of verification mechanisms in high-stakes application and diligent examination from different linguistic backgrounds.

\subsection{Failure Case Analysis and Model Limitations}

Comprehensive analysis of model failures provides critical insights for understanding limitations and guiding future improvements, particularly important for sensitive applications requiring high reliability.

\subsubsection{Hinglish Model Failure Patterns}

The code-mixed nature of Hinglish presents unique challenges that extend beyond those observed in monolingual detection systems, requiring specialized analysis to understand multilingual and multicultural complexities.

\textbf{Color Coding Legend:}
\begin{itemize}
    \item \textcolor{red}{Red}: Offensive or aggressive language.
    \item \textcolor{orange}{Orange}: Subtle discriminatory or biased terms.
    \item \textcolor{green}{Green}: Positive words used sarcastically or ironically.
\end{itemize}

\begin{table}[h]
\centering
\caption{Failure case analysis for Hinglish cyberbullying detection}
\label{tab:hinglish_failure_cases}
\small
\setlength{\tabcolsep}{4pt}
\begin{tabular}{|c|p{6.2cm}|c|p{0.9cm}|p{4.2cm}|}
\hline
\textbf{S.No.} & \textbf{Text} & \textbf{Truth} & \textbf{Pred.} & \textbf{Analysis} \\
\hline
1 & ``Kismat Badi \textcolor{red}{kutti} cheez hain... \textcolor{red}{Saali} Palat hi nht rahi." (Luck is such a \textcolor{red}{damned} thing...it's not turning around.) & \begin{tabular}[c]{@{}c@{}}Non\\bully\end{tabular} & \begin{tabular}[c]{@{}c@{}}Bully\end{tabular} & Misinterpreted vulgar terms directed at abstract concept (luck) rather than a person. \\
\hline
2 & ``Tere jaise \textcolor{orange}{bewakoof} log hi is desh ko barbaad kar rahe hain." (\textcolor{orange}{Foolish} people like you are ruining this country.) & \begin{tabular}[c]{@{}c@{}}Bully\end{tabular} & \begin{tabular}[c]{@{}c@{}}Non\\bully\end{tabular} & Missed subtle personal attack disguised as political commentary. Cultural context crucial for detection. \\
\hline
3 & ``Yaar tu bohot \textcolor{green}{smart} hai, bilkul Einstein jaise!" (Dude, you're so \textcolor{green}{smart}, just like Einstein!) & \begin{tabular}[c]{@{}c@{}}Bully\end{tabular} & \begin{tabular}[c]{@{}c@{}}Non\\bully\end{tabular} & Failed to detect sarcastic praise as implicit mockery. Irony difficult across languages. \\
\hline
4 & ``\textcolor{red}{BC} kya \textcolor{red}{chutiya} bana raha hai yeh banda." (What a \textcolor{red}{fool} this guy is making of himself.) & \begin{tabular}[c]{@{}c@{}}Bully\end{tabular} & \begin{tabular}[c]{@{}c@{}}Non\\bully\end{tabular} & Offensive language with personal targeting missed due to informal expression patterns. \\
\hline
\end{tabular}
\end{table}

The Hinglish model exhibits distinct failure patterns reflecting the complexity of code-mixed communication:

\textbf{Context-Insensitive Lexical Detection:} The model occasionally classifies hateful words based on how they are generally used and overlooks instances where hateful words are employed for defense or non-targeting contexts. The trend indicates that there is a need for better contextual understanding mechanisms.

\textbf{Target Ambiguity:} It is difficult to distinguish between attacks on an individual and criticism of abstract entities, especially when offensive terms are used metaphorically. This indicates the requirement for stronger target identification skills.

\textbf{Cultural Subtext:} Having deep roots in personal attacks hidden deep within political or cultural remarks require profound cultural insight and much more than the present model constraints. This highlights the necessity of incorporating cultural insight into model training.

\textbf{Cross-linguistic Sarcasm:} Sarcastic sentences that cross linguistic boundaries are harder to resolve because ironic meaning is less easily identified in the presence of other linguistic codes. This suggests a need for more cross-linguistic sentiment analysis competence.

\textbf{Informal Expression Patterns:} Extremely informal or extremely slang-rich sentences often go unnoticed because their non-standard form. This indicates that more variation in informal sentence structures has to be incorporated into training sets.

These patterns of failure offer important lessons for future model development and point to challenges still to be met in automated detection of cyberbullying on code-mixed platforms. The analysis suggests the necessity for ongoing research in context-sensitive and culturally-aware detection mechanisms for multilingual social media.

\section{Conclusion and Future Work}

This work effectively fills the vital gap between correct cyberbullying identification and explainable AI in Hindi-English code-mixed settings through building a robust MURIL-based model that outperforms state-of-the-art multilingual models on a diverse range of benchmarking datasets with responsible content moderation using transparent decision-making mechanisms. The suggested method illustrates the spectacular benefits of task-guided multilingual pretraining for code-mixed data processing, attaining spectacular performance gains of 2.52 to 13.07 percentage points against rival architectures by using advanced management of cross-lingual phenomena, cultural references, and transliteration words typical of Hinglish conversation. The five-data-set overall comprehensive evaluation, careful ablation tests unveiling the best-performing architectural setups, and strict explainability evaluation via word attribution and cross-linguistic pattern discovery all set new standards in multilingual cyberbullying detection while fulfilling deployment needs required in real-world scenarios. The ability of the framework to deal with complex code-mixing situations to high accuracy and interpretability is a big leap towards more culturally-sensitive and inclusive content moderation systems for multilingual online interactions among multicultural communities.

Subsequent studies will address overcoming the above limitations and broadening the scope of the framework to even more multilingual populations and emerging social media spaces. The development of culturally-aware detection mechanisms that can better understand subtle personal attacks embedded within cultural or political commentary represents a critical advancement opportunity, potentially through the integration of cultural knowledge graphs and context-aware training methodologies that capture region-specific social dynamics and linguistic patterns. Additionally, investigating the integration of conversational context, user behavioral patterns, and temporal dynamics could enhance detection accuracy for subtle forms of cyberbullying that manifest across multiple interactions rather than single messages, while the development of proactive intervention mechanisms based on early warning signals could transform the framework from reactive detection to preventive protection in multilingual social media environments.

\end{document}